\documentclass[conference]{IEEEtran}
\IEEEoverridecommandlockouts
\usepackage{cite}
\usepackage{amsmath,amssymb,amsfonts}
\usepackage{algorithmic}
\usepackage{graphicx}
\usepackage{textcomp}
\usepackage{xcolor}
\usepackage{verbatim}
\usepackage{booktabs}
\usepackage{multirow}
\usepackage{diagbox}
\usepackage{url}
\usepackage{array}
\usepackage{lipsum}
\begin{document}

\title{RVAE-LAMOL: Residual Variational Autoencoder to Enhance Lifelong Language Learning\\
}
\author{
\IEEEauthorblockN{Han Wang\textsuperscript{1,2}, Ruiliu Fu\textsuperscript{1,2}, Xuejun Zhang\textsuperscript{1,2}, Jun Zhou\textsuperscript{1,2}}
\IEEEauthorblockA{\textsuperscript{1}\textit{Key Laboratory of Speech Acoustics and Content Understanding} \\
{\textit{Institute of Acoustics, Chinese Academy of Sciences, China}} \\
\textsuperscript{2}\textit{University of Chinese Academy of Sciences, Beijing, China}\\
\{wanghan, furuiliu, zhangxuejun, zhoujun\}@hccl.ioa.ac.cn}
}

\maketitle

\newcommand\blfootnote[1]{%
\begingroup
\renewcommand\thefootnote{}\footnote{#1}%
\addtocounter{footnote}{-1}%
\endgroup
}

\blfootnote{*This paper has been accepted for publication at IJCNN 2022 on IEEE WCCI 2022.} 

\begin{abstract}
Lifelong Language Learning (LLL) aims to train a neural network to learn a stream of NLP tasks while retaining knowledge from previous tasks. However, previous works which followed data-free constraint still suffer from catastrophic forgetting issue, where the model forgets what it just learned from previous tasks. In order to alleviate catastrophic forgetting, we propose the residual variational autoencoder (RVAE) to enhance LAMOL, a recent LLL model, by mapping different tasks into a limited unified semantic space. In this space, previous tasks are easy to be correct to their own distribution by pseudo samples. Furthermore, we propose an identity task to make the model is discriminative to recognize the sample belonging to which task. For training RVAE-LAMOL better, we propose a novel training scheme Alternate Lag Training. In the experiments, we test RVAE-LAMOL on permutations of three datasets from DecaNLP. The experimental results demonstrate that RVAE-LAMOL outperforms naïve LAMOL on all permutations and generates more meaningful pseudo-samples.
\end{abstract}

\begin{IEEEkeywords}
Residual VAE, lifelong language model learning, GPT-2
\end{IEEEkeywords}

\section{Introduction}
Lifelong learning is the ability of humans \cite{Ring97} that gain, strengthen, and transfer knowledge during their lifetime. The human can learn new knowledge and consolidate old knowledge through building internal connections between new and old knowledge by simplifying the complexity and breaking it down one by one. It is also important for natural language processing (NLP) to learn various kinds of NLP tasks like humans. For instance, during the applying application of NLP, a lot of new data is continuously acquired, including new data of existing requirements and new requirements. For new data with existing requirements, the traditional method, known as isolated learning \cite{lll-book-2nd}, is to extend previous training data with them and train the model from scratch. For new requirements, multi-task learning applies the new data to be jointly trained with previous requirements training data to optimize the model from scratch. Both isolated learning and multi-task learning train the model with the limited assumption that obtaining all data during training. When a stream of tasks are trained sequentially, neural network models suffer from catastrophic forgetting \cite{Ring97,mccloskey1989catastrophic,french1999catastrophic} that the model forgets the early learned tasks.

Lifelong language learning (LLL), which is focused in this paper, aims to learn a stream of NLP tasks via lifelong learning. Recently, LAMOL \cite{sun2019lamol} is a general generative LLL framework that uses a language model (LM) to learn various kinds of NLP tasks in question answering (QA) format. Instead of using real data from previous tasks, LAMOL generates pseudo data for previous tasks and then uses these pseudo data to joint train with a new task. During inference, results can be generated by inputting the contents and questions of samples. LAMOL's simplicity and effectiveness have recently attracted a lot of attention. However, LAMOL's performance still falls short of what is usually thought to be the upper bound of LLL's performance: multi-task learning.
This indicates that only naive generating pseudo samples maybe not be sufficient to alleviate catastrophic forgetting.

We find that the pseudo samples (see Table \ref{tab:pseudo samples} in Appendix) generated by LAMOL do not always correspond to their own task-specific tokens especially becoming worse when fewer pseudo samples for previous tasks are generated and jointly trained with the new task. This indicates that the naive task-specific token is not enough to restrict the semantic space of the task from being affected by the change of data distribution. Once the early learned tasks are biased by the new task, it is hard to be correct them with a few pseudo samples. 

Therefore, in this paper, we propose RVAE-LAMOL, which contains the identity task and the residual variational autoencoder (RVAE), to enhance LAMOL. Firstly, we introduce the identity task, which identifies which task the content of the sample belongs to, to improve the correlation between the sample's content and its own task-specific token. Secondly, we introduce RVAE to reconstruct the internal hidden states of LAMOL. In this way, RVAE can map different tasks into a limited unified semantic space that the biased tasks can be corrected to their own semantic space by a few pseudo-data. Therefore, more pseudo samples are generated corresponding to their own task-specific generation tokens. For more effective training, we propose Alternate Lag Training (ALT) to split each task training into multiple turns. RVAE is only trained at the second half of each turn. In this way, RVAE is always trained on the fine-tuned GPT-2 \cite{gpt2} resulting in better performance.

\begin{figure}[ht]
\centering 
\includegraphics[scale=0.3]{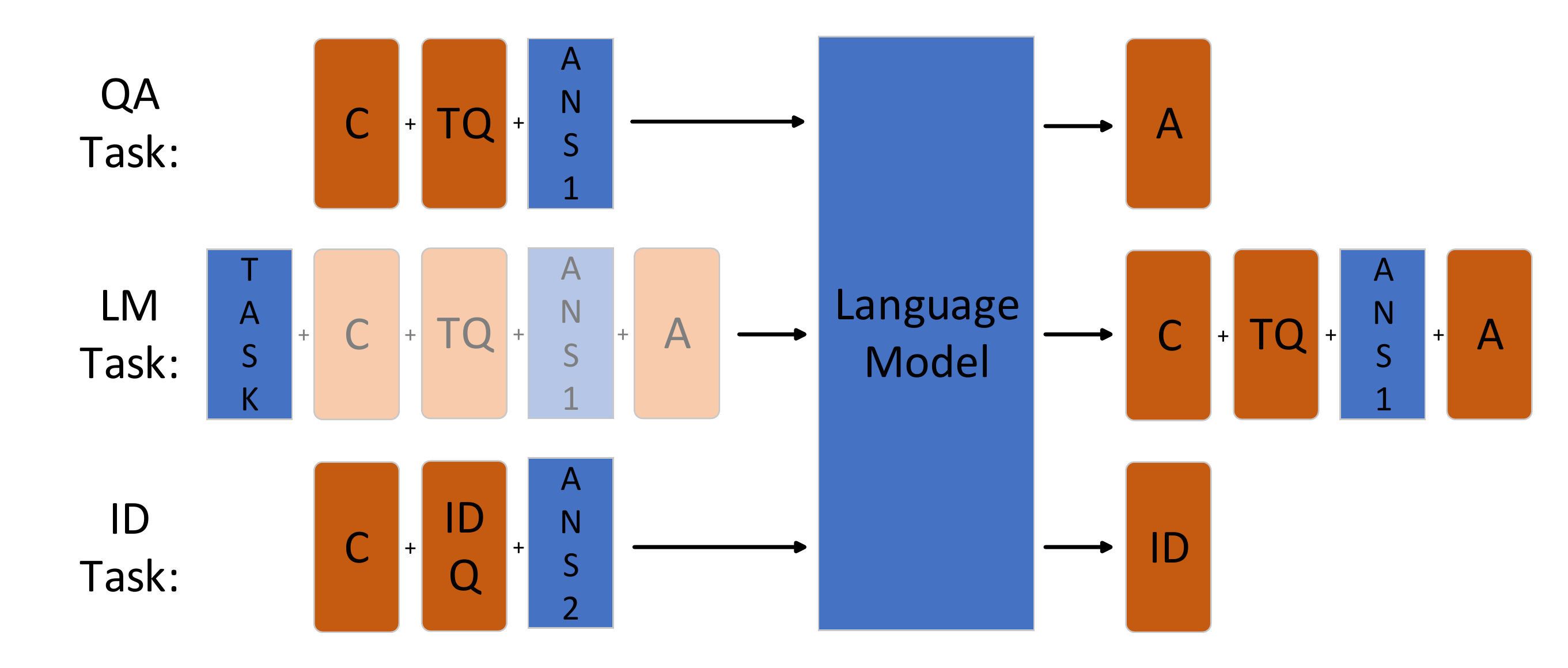} 
\caption{Illustration of QA/LM/ID task.}
\label{all tasks}
\end{figure}

The contributions of our paper are listed below:
\begin{itemize}
    \item We demonstrate the effectiveness of the identity task and RVAE when pseudo-samples are sufficient and insufficient. Both of them can alleviate catastrophic forgetting by improving the correlation between the sample's content and its own task-specific token. 
    \item We propose a new training scheme ALT to train RVAE-LAMOL better by splitting a difficult training process into multiple easy turns. RVAE is always trained based on a nice baseline. In this way, our RVAE-LAMOL is more robust to the position of RVAE in GPT-2.
    \item We expand RVAE to residual conditional variational autoencoder (RCVAE) by observing the task identity information, the task-specific token, to improve the performance when fewer pseudo-data. 
\end{itemize}

We evaluate RVAE-LAMOL\footnote{\url{https://github.com/CodeHan/RVAE-LAMOL}} on six task order permutations of three datasets from DecaNLP \cite{decaNLP} following LAMOL. The experimental results demonstrate that our proposed RVAE-LAMOL outperformed the original LAMOL on all permutations, achieving average improvements of 2.1 and 5.8 percentage points when sufficient and insufficient pseudo-data, respectively. Moreover, observing the task identity information when insufficient pseudo-data also yielded competitive performance, achieving average improvements of 3.2 percentage points from RVAE-LAMOL.

\section{Related Work}
\subsection{VAE}
Variational Autoencoder (VAE) \cite{vae2014-kingma} is a latent variable model that approximate a data distribution $p(X)$. VAE first encodes input x into mean value $\mu$ and standard deviation $\sigma$. Then, the latent feature $z$ is obtained by reparametrization trick with $\mu$ and $\sigma$. Finally, VAE produce $x$ by a decoder $p(x|z)$. The objective of VAE is to maximize the marginal likelihood $logp(x)$ by optimizing the evidence lower bound (ELBO):
\begin{equation}
\mathcal{L}_{VAE}(\theta,\phi) = \mathbb{E}_{z \sim q_{\phi}(z|x)}[logp_{\theta}(x|z)]-KL(q_{\phi}(z|x)||p(z)) \label{eq_naive_vae}
\end{equation}
where $\phi$ and $\theta$ denote the inference network (encoder) and the decoder, respectively. In \eqref{eq_naive_vae}, the first term is reconstruction loss and the second one is the KL divergence between the approximate posterior and the prior. Generally, the Gaussian distribution $\mathcal{N} \sim (0,I)$ is choice for the prior. Then, the KL divergence can be computed as:
\begin{equation}
KL=\frac{1}{2}\sum_{i=1}^{n}(\mu_{i}^{2}+\sigma_{i}^{2}-log\sigma_{i}^{2}-1)    \label{eq_vae_kl}
\end{equation}
where $\mu_{i}$, $\sigma_{i}$ denotes the $i$-th dimension of the mean and standard deviation.

Conditional VAE (CVAE) \cite{cvae-2015nips,cvae-2017acl} is a adaptation of VAE to fit supervised learning and conditional generation. Given an observation $y$, CVAE maximize the conditional marginal likelihood $logp(x|y)$ by optimizing the evidence lower bound (ELBO):

\begin{equation}
\begin{aligned}
&\mathcal{L}_{VAE}(\theta,\phi) = &\mathbb{E}_{z \sim q_{\phi}(z|y,x)}[logp_{\theta}(x|y,z)]\\
& &-KL(q_{\phi}(z|y,x)||p(z|y)) \label{eq_naive_cvae}
\end{aligned}
\end{equation}
Recently, several works proposed utilized Transformers-based VAE to complete various NLP tasks: machine translation \cite{VTN-2019EMNLP-MT,VTN-2020-Disentangled-Context-MT,VTN-2020AAAI-machine-translation}, sentiment analysis \cite{VTN-2019CoNLL-SA}, dialogue generation \cite{VTN-2020-response-generation} and story generation \cite{VTN-2019IJCAI-story-completion,VTN-2019IJCNN-sentence-generation,VTN-2021-controllable-story-generation}. These works build or reconstruct the latent feature by Transformers. Inspired by these works, Transformers is a nice architecture for VAE to generate controllable text. 

In this paper, we apply Linear-based residual VAE to map the hidden states of Transformers into a limited semantic space for generating controllable text, instead of Transformers-based VAE.

\subsection{Lifelong Language Learning}
Lifelong language learning (LLL) is an essential step in promoting the realization of general artificial intelligence in the field of NLP. 
\cite{lll-2015ACL-SC,lll-2019NAACL-SR} have studied LLL on a single type of NLP task. MbPA++ \cite{d2019mbpa} has studied text classification and question answering by using episodic memory to preserve some real samples of previous tasks. The episodic memory is replayed in the training process and used to local adaptation on inference. Meta-MbPA \cite{meta-mbpa} applies a meta-lifelong framework to improve MbPA++. Recently, LAMOL \cite{sun2019lamol} uses a language model (LM) to learn various kinds of NLP tasks in QA-style. In LAMOL, the pseudo-data generated by the model is trained together with the new task to alleviate catastrophic forgetting. L2KD \cite{l2kd} and DnR \cite{dnr} distilled parts of GPT-2 \cite{gpt2} layers to improve LAMOL. Rational-LAMOL \cite{rational-lamol} applied critical freezing guided by rationale information which is obtained by human or unsupervised rationale generation \cite{2020ICML-unsupervise-rationale}.

In contrast to previous work, our proposed RVAE-LAMOL is not only a data-based LLL method like previous LAMOL-based works but also an architecture-based method that applies residual VAE to map various tasks into a limited unified but task-discriminative semantic space. Hence, the model can generate more reasonable pseudo-samples to alleviate catastrophic forgetting.

\section{Methodology}\label{Methodology}

\begin{figure}[htbp]
\centering 
\includegraphics[scale=0.3]{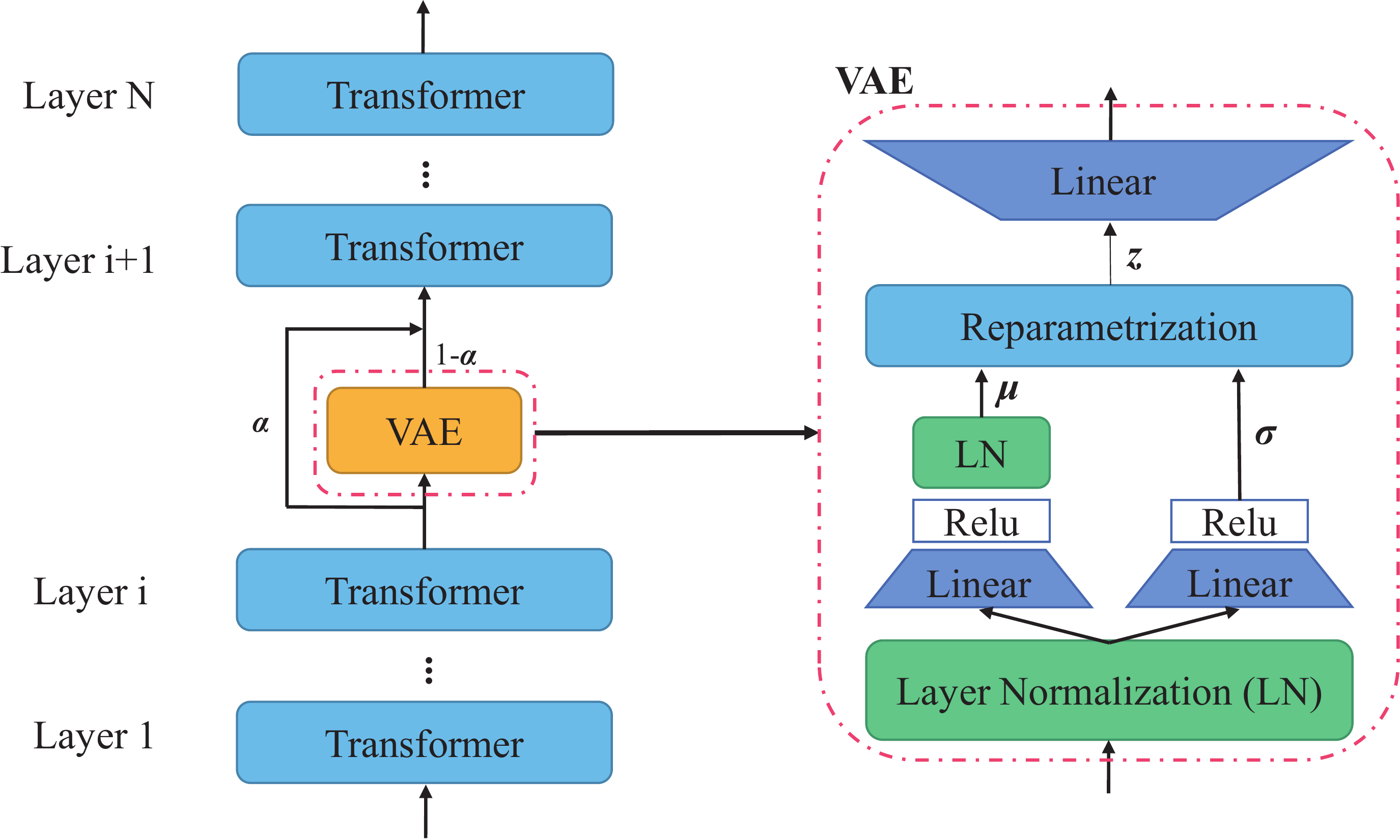} 
\caption{The framework of RVAE-LAMOL.}
\label{framework}
\end{figure}

\subsection{LAMOL}\label{section-lamol}
Language Modeling for Lifelong Language Learning (LAMOL) \cite{sun2019lamol} is a lifelong learning framework to use a single language model (LM) to learn different NLP tasks in sequence. In LAMOL, various kinds of NLP tasks are formatted into question-answering (QA) style and fed into GPT-2 to learn two tasks: LM task and QA task. Each sample contains five parts: a task-specific generation token [TASK]/[GEN], context (C), task question (TQ), answer token [ANS], and answer (A). LM task aims to train GPT-2 as a generative model by receiving [TASK]/[GEN] and then generating C+TQ+[ANS]+A. QA task is to train GPT-2 can generate A by giving C+TQ+[ANS]. Both LM and QA task are illustrated as Fig. \ref{all tasks}. The loss of LAMOL is as follows:
\begin{equation}
\mathcal{L}_{base} = \mathcal{L}_{QA}+\lambda \mathcal{L}_{LM}
\end{equation}
where $\lambda$ is the weight of the LM task.

Before training a new task, LAMOL can generate pseudo samples for previously learned tasks. Assuming that $T = \{T_{1},T_{2},…,T_{N}\}$ represents N tasks to be learned. Let $\gamma$ and $|D_{t}|, t \in [2, N]$ denote the sampling ratio and size of $t$-th task, respectively. LAMOL generates $\frac{\gamma}{t-1}|D_{t}|$ pseudo-samples for each previously learned task. All pseudo-samples are joint trained with the new task and alleviate the long-standing issue of lifelong learning–catastrophic forgetting. However, not each pseudo sample is tied to [TASK]/[GEN] resulting in the gap between LAMOL and multi-task learning (MTL). In this paper, we propose the identity task and residual VAE for LAMOL to improve the tightness of the pseudo sample and the task-specific generation token, discussed next.

\subsection{Identity Task}\label{section-id task}
The identity (ID) task is proposed to tackles the confusion problem, in which the content of the pseudo sample is not corresponding to the task-specific generation token when generating pseudo samples. As described in Section \ref{section-lamol}, we can know that the ratio between the size of the new task and each previous task is $R=\frac{t-1}{\gamma}$. The more learned tasks are and the lower sampling ratio $\gamma$ is, the larger is $R$. The larger $R$ means that a more imbalanced dataset is learned by the model which is confusing to identify the mapping relationship between the content of the pseudo sample and the task-specific generation token. This confusion problem makes the model unstable when generating pseudo samples. The ID task, a simple yield efficient task, makes the model can identify which task the C of the sample belongs to.
	
ID task format the sample into four parts: context (C), ID question (IDQ), ID answer token [ANS2], task-specific generation token [TASK]. ID task is illustrated in Fig. \ref{all tasks}. ID task is to generate [TASK] by giving C+IDQ+[ANS2]. The loss can be updated as follows:
\begin{equation}
\mathcal{L} = \mathcal{L}_{base}+\beta \mathcal{L}_{ID} \label{eq_all_task_loss}
\end{equation}
where $\beta$ is the weight of the ID task.

\subsection{Residual VAE-LAMOL}\label{setion-rvae}
We propose residual VAE-LAMOL (RVAE-LAMOL) to alleviate the confusion among different tasks semantic space. As described in Section \ref{section-lamol} and \ref{section-id task}, not each pseudo sample is tied to the task-specific generation token resulting in the gap between LAMOL and MTL. The reason for this gap is as follows: GPT-2 can learn the LM task through fine-tuning, but it does not constrain each task in semantic space. Different tasks are far apart in semantic space. No matter how small $R$ is, the previous tasks are biased by the new task in semantic space, resulting in confusing samples, which eventually lead to catastrophic forgetting.

VAE can encode and decode features of different tasks and uses the learned latent feature to unify different tasks into latent semantic space. VAE encode input feature of tasks to obtain mean value $\mu$ and standard deviation $\sigma$. Then, the latent feature $z$ is generated by reparametrization trick with $\mu$ and $\sigma$ (see as \eqref{eq_z}, \eqref{eq_mu}, \eqref{eq_sigma}). Finally, $z$ is decoded to a new feature which is constrained to close to the input feature. Training VAE to obtain $z$ can map different tasks into a limited unified semantic space. In this space, different tasks distribute discriminatively but not much far away from each other. Therefore, previous tasks are easy to be correct to their own distribution in latent semantic space by a few pseudo samples, resulting to generate pseudo samples corresponding to the task-specific generation token [TASK].

\begin{align}
    RVAE(h_{i}) &=  (1-\alpha)Decoder(z)+\alpha h_{i} \label{eq_rvae}\\
    z &= Reparametrization(\mu,\sigma) \label{eq_z}\\
    \mu &= LN(Relu(Encoder_{\mu}(LN(h_{i})))) \label{eq_mu}\\
    \sigma &= Relu(Encoder_{\sigma}(LN(h_{i}))) \label{eq_sigma}
\end{align}

\subsubsection{Architecture} In this paper, We propose weighted residual VAE instead of naive VAE, which makes transformer and VAE complement each other. On the one hand, VAE can map different tasks into a limited unified semantic space, but it is hard to accurately reconstruct the complex feature of Transformers, which makes it difficult for the QA/LM/ID task to converge to the optimum. On the other hand, VAE is sub-optimized when there is a naive residual connection \cite{resnet-2016CVPR}. Therefore, we apply weighted residual connections to resolve the two mentioned problems. As shown in \eqref{eq_rvae}, $\alpha$ is the factor of RVAE. 
\begin{figure}[htp]
\centering 
\includegraphics[scale=0.7]{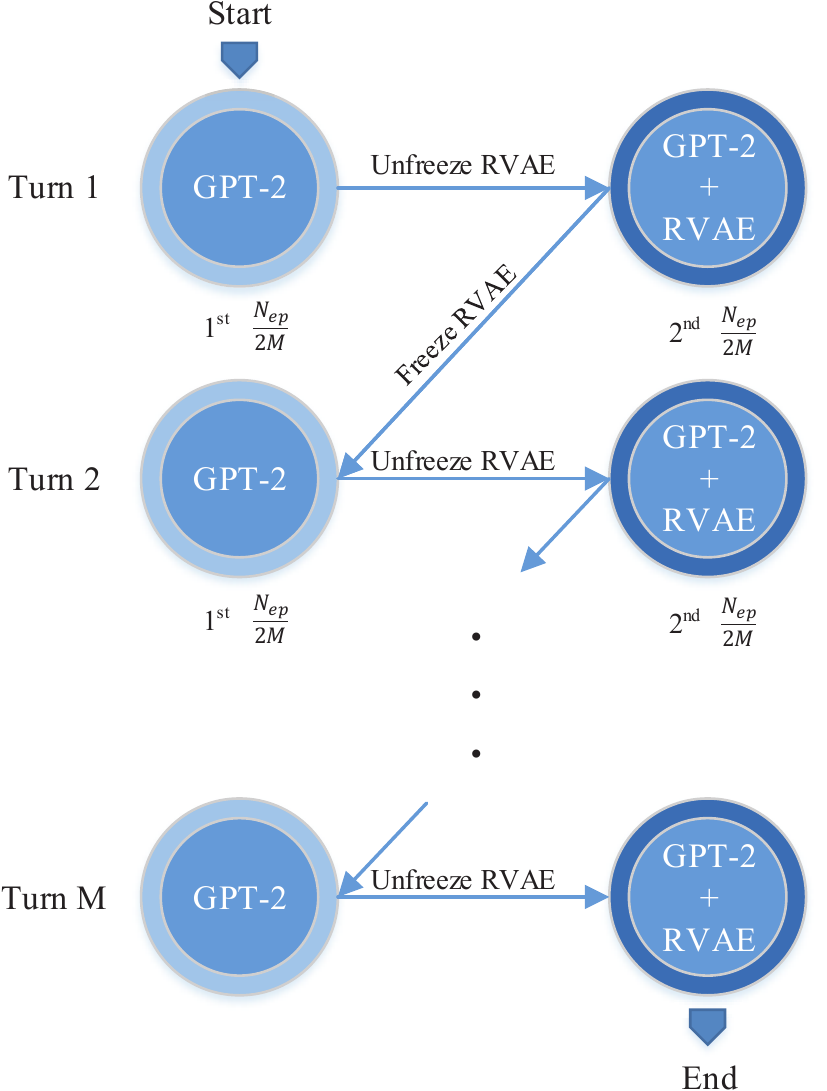} 
\caption{Illustration of Alternate Lag Training.}
\label{alt}
\end{figure}
Furthermore, \cite{bn-vae-2020ACL} proposed to apply Batch Normalization (BN) to prevent posterior collapse of VAE. In our proposed RVAE, we apply Layer Normalization (LN) instead of BN due to the large difference in the length of samples (see as \eqref{eq_mu}). However, LN is not enough to prevent posterior collapse when the architecture of VAE is simple. \cite{vae-fb-2016NIPS,vae-fb-2019EMNLP} proposed Free Bits (FB) to replace KL term in ELBO with a hinge loss term:
\begin{equation}
\sum_{i}max(\rho,KL(q_{\phi}(z_{i}|x)||p(z_{i})))\label{eq_vae_kl_fb}
\end{equation}
where $\rho$ denotes the target rate and $z_i$ denotes the $i$-th dimension of the latent feature z. The experiments of \cite{vae-fb-2019EMNLP} demonstrated that FB can make VAE achieve better performance in terms of both language modeling and reconstruction. Therefore, we utilize FB together with LN to prevent RVAE from posterior collapse.

The framework of our proposed RVAE-LAMOL is illustrated in Fig. \ref{framework}.

\subsubsection{Training Process}
To train RVAE-LAMOL better and more robust, we propose Alternate Lag Training (ALT). The general training process is that jointly optimizes all components of the model during whole epochs. However, RVAE-LAMOL is composed of pretrained GPT-2 and randomly initialized RVAE. GPT-2 is harder to be trained than RVAE because of the complex architecture and huge parameters of the transformer. Therefore, in ALT, we split all epochs $N_{ep}$ into $M$ turns. Each turn contains $\frac{N_{ep}}{M}$ epochs. In the first half epochs of each turn, we only train GPT-2, and the remaining epochs train both GPT-2 and RVAE. Since ALT enables RVAE to be trained on a better baseline,  RVAE-LAMOL can map different tasks into a more reasonable semantic space. M should be greater than 1. If $M=1$, the ALT degenerates into the general training process, resulting in RVAE-LAMOL being suboptimal. ALT is illustrated in Fig. \ref{alt}.

\begin{table}[ht]
\caption{Datasets sizes and evaluation metrics.}
\centering
\begin{tabular}{@{}llll@{}}
\toprule
\textbf{Dataset} & \textbf{\#Train} & \textbf{\#Test} & \textbf{Metric} \\ \midrule
WOZ              & 2536           & 1646          & dsEM    \\
QA-SRL           & 6414           & 2201          & nF1     \\
SST              & 6920           & 1821          & EM      \\  \bottomrule
\end{tabular}
\label{datasets}
\end{table}

\subsubsection{Extension to CVAE}
Further to make the model generate more corresponding pseudo data, we extend our RVAE to Residual Conditional Variational Autoencoder (RCVAE)  by observing task identity information during the LM task and pseudo-data generation. In \eqref{eq_naive_cvae}, the observation y is the task-specific token [TASK].

\subsection{Overall Training Objective}
Our proposed RVAE-LAMOL, which is introduced in previous section, is applied to the QA/LM/ID task mentioned in Section \ref{section-lamol} and \ref{section-id task}. Hence, the loss of each task in \eqref{eq_all_task_loss} is updated as follows:

\begin{equation}
    \mathcal{L}_{task} = \mathcal{L}_{task}^{naive}+\mathcal{L}_{task}^{VAE} \label{eq_task_loss_vae}
\end{equation}
where $task$ is the placeholder of QA/LM/ID task. In Eq. \eqref{eq_task_loss_vae}, the first term is the naive objective of the language model and the second term is the VAE objective with FB (see as \eqref{eq_naive_vae}, \eqref{eq_vae_kl} and \eqref{eq_vae_kl_fb}).

\section{Experiment Setup}
To evaluate our proposed RVAE-LAMOL, we conduct a set of experiments detailed below.
\subsection{Implementation Detail}
To make the results comparable, we build our RVAE-LAMOL based on the implementation of LAMOL\footnote{\url{https://github.com/jojotenya/LAMOL}}. We also use pretrained small GPT-2 \cite{gpt2} for all tasks. We use the best settings in LAMOL: using the task-specific token as the beginning token of pseudo-samples and the weight of LM task $\lambda =0.25$ and the optimizer is AdamW \cite{AdamW-2017}. During inference, we use greedy decoding to generate sequence. The weight of ID task $\beta =0.5$. We use single Linear layer for the encoder and decoder of RVAE. The dimension of latent feature $z$ is 100. RVAE is located behind the second layer of GPT-2. The factor of RVAE $\alpha =0.5$ and the target rate $\rho =0.2$. We train each tasks 24 epochs and set the ALT turns $M=3$. The learning rate is 1e-4. We use a single NVIDIA TESLA P100 (12GB) for each experiment.

\begin{table*}[htbp]
  \centering
  \caption{Experiments on three DecaNLP tasks: [SST, QA-SRL (SRL), WOZ].}
    \begin{tabular}{c|c|cccccccc}
    \toprule
    $\gamma$ & Methods & \scriptsize{SST SRL WOZ} & \scriptsize{SST WOZ SRL} & \scriptsize{SRL SST WOZ} & \scriptsize{SRL WOZ SST} & \scriptsize{WOZ SST SRL} & \multicolumn{1}{c|}{\scriptsize{WOZ SRL SST}} & \multicolumn{1}{c|}{Average} & Std. \\
    \midrule
    \multirow{6}[2]{*}{0.01} & LAMOL & 73.7  & 72.2  & 64.2  & 70.1  & 51.9  & \multicolumn{1}{c|}{60.9} & \multicolumn{1}{c|}{65.5} & 7.5 \\
          & LAMOL+ID task & 77.5  & 73.0    & 66.9  & 71.6  & 53.4  & \multicolumn{1}{c|}{62.2} & \multicolumn{1}{c|}{67.4} & 7.9 \\
          & RVAE-LAMOL & 77.8  & 69.5  & 65.6  & 72.9  & 53.5  & \multicolumn{1}{c|}{59.5} & \multicolumn{1}{c|}{66.5} & 8.1 \\
          & \multicolumn{1}{r|}{w/o ID task} & 76.6  & \textbf{75.9} & \textbf{70.7} & 71.6  & \textbf{61.2} & \multicolumn{1}{c|}{63.1} & \multicolumn{1}{c|}{69.9} & 5.9 \\
          & RCVAE-LAMOL & \textbf{78.5} & 73.5  & 68.8  & 72.5  & 54.1  & \multicolumn{1}{c|}{\textbf{71.0}} & \multicolumn{1}{c|}{69.7} & 7.6 \\
          & \multicolumn{1}{r|}{w/o ID task} & 76.9  & 74.8  & 69.2  & \textbf{73.5} & 59.3  & \multicolumn{1}{c|}{69.9} & \multicolumn{1}{c|}{\textbf{70.6}} & \textbf{5.7} \\
    \midrule
    \multirow{6}[2]{*}{0.03} & LAMOL & 76.0    & 74.8  & 71.6  & 74.3  & 57.5  & \multicolumn{1}{c|}{70.6} & \multicolumn{1}{c|}{70.8} & 6.2 \\
          & LAMOL+ID task & \textbf{79.7} & 75.0    & 76.8  & 74.3  & 68.8  & \multicolumn{1}{c|}{71.4} & \multicolumn{1}{c|}{74.3} & 3.5 \\
          & RVAE-LAMOL & 79.0    & 74.8  & \textbf{78.5} & 75.7  & \textbf{75.3} & \multicolumn{1}{c|}{76.1} & \multicolumn{1}{c|}{\textbf{76.6}} & \textbf{1.6} \\
          & \multicolumn{1}{r|}{w/o ID task} & 79.2  & \textbf{79.3} & 77.4  & \textbf{77.3} & 69.5  & \multicolumn{1}{c|}{75.9} & \multicolumn{1}{c|}{76.4} & 3.3 \\
          & RCVAE-LAMOL & 79.6  & 78.2  & 77.3  & 75.8  & 70.8  & \multicolumn{1}{c|}{\textbf{77.8}} & \multicolumn{1}{c|}{\textbf{76.6}} & 2.8 \\
          & \multicolumn{1}{r|}{w/o ID task} & 73.5  & 79.0    & 77.4  & 76.1  & 73.2  & \multicolumn{1}{c|}{76.6} & \multicolumn{1}{c|}{76.0} & 2.1 \\
    \midrule
    \multirow{6}[2]{*}{0.05} & LAMOL & 77.3  & 76.9  & 78.1  & 74.7  & 73.4  & \multicolumn{1}{c|}{75.8} & \multicolumn{1}{c|}{76.0} & 1.6 \\
          & LAMOL+ID task & 79.0    & 76.6  & 76.8  & 76.7  & 76.2  & \multicolumn{1}{c|}{77.8} & \multicolumn{1}{c|}{77.2} & 0.9 \\
          & RVAE-LAMOL & 79.6  & 78.9  & 78.8  & 76.9  & 78.4  & \multicolumn{1}{c|}{76.4} & \multicolumn{1}{c|}{78.2} & 1.1 \\
          & \multicolumn{1}{r|}{w/o ID task} & 79.5  & 78.1  & \textbf{79.1} & \textbf{78.7} & 77.8  & \multicolumn{1}{c|}{77.5} & \multicolumn{1}{c|}{78.5} & \textbf{0.7} \\
          & RCVAE-LAMOL & \textbf{81.2} & \textbf{79.2} & 79.0    & 77.6  & \textbf{78.8} & \multicolumn{1}{c|}{\textbf{78.2}} & \multicolumn{1}{c|}{\textbf{79.0}} & 1.1 \\
          & \multicolumn{1}{r|}{w/o ID task} & 79.6  & \textbf{79.2} & 79.0    & 78.2  & 77.6  & \multicolumn{1}{c|}{79.9} & \multicolumn{1}{c|}{78.9} & 0.8 \\
    \midrule
    \multirow{6}[2]{*}{0.2} & LAMOL & 79.4  & 79.9  & 80.1  & 78.7  & 79.8  & \multicolumn{1}{c|}{79.0} & \multicolumn{1}{c|}{79.5} & 0.5 \\
          & LAMOL+ID task & \textbf{81.7} & 80.3  & 80.6  & 80.3  & 80.7  & \multicolumn{1}{c|}{79.8} & \multicolumn{1}{c|}{80.6} & 0.6 \\
          & RVAE-LAMOL & 81.3  & \textbf{82.4} & \textbf{81.2} & 81.6  & \textbf{81.5} & \multicolumn{1}{c|}{\textbf{81.8}} & \multicolumn{1}{c|}{\textbf{81.6}} & 0.4 \\
          & \multicolumn{1}{r|}{w/o ID task} & 81.1  & 81.6  & 80.7  & 80.8  & 81.0    & \multicolumn{1}{c|}{81.4} & \multicolumn{1}{c|}{81.1} & \textbf{0.3} \\
          & RCVAE-LAMOL & 80.9  & 81.2  & 80.3  & \textbf{81.9} & 81.2  & \multicolumn{1}{c|}{81.0} & \multicolumn{1}{c|}{81.1} & 0.5 \\
          & \multicolumn{1}{r|}{w/o ID task} & 80.9  & 81.9  & 80.5  & 81.0    & 80.4  & \multicolumn{1}{c|}{80.9} & \multicolumn{1}{c|}{80.9} & 0.5 \\
    \midrule
    /     & Multi-task & \multicolumn{8}{c}{81.5} \\
    \bottomrule
    \end{tabular}%
  \label{table-deca3}%
\end{table*}%

\subsection{Datasets}\label{section-dataset}
To evaluate the capability of RVAE-LAMOL on diverse sequence generation tasks, we pick the following three tasks from DecaNLP \cite{decaNLP} following LAMOL: English Wizard of Oz (WOZ) \cite{wen2016woz} is a dataset of oriented dialogue of the restaurant reservation task; QA-SRL \cite{he2017srl} is a dataset of semantic role labeling in SQuAD-style; SST is a binary version of The Stanford Sentiment Treebank \cite{radford2017sst} with positive and negative sentiment. These datasets are three different NLP tasks that are trained in random order to evaluate whether our proposed method is robust to a variety of NLP tasks. A normalized F1 (nF1) metric that lowers text and removes punctuation and articles, is used to evaluate SRL. The exact match (EM) is used to evaluate SST. The turn-based dialogue state EM (dsEM) is used for WOZ. The information about datasets is shown in Table \ref{datasets}.

\section{Results and Discussion}
In this section, we firstly conduct several experiments to report the performance of RVAE-LAMOL and compare it with LAMOL as the baseline as well as multitask learning, which is considered as the upper bound of LL. Secondly, we validate the effectiveness of the ALT. Thirdly, we analyze the effectiveness of the ID task. Finally, we analyze the effectiveness of RVAE in terms of its position in GPT-2 and the dimension of the latent feature.

\subsection{Results on Three DecaNLP Tasks}\label{section deca3 result}
To validate our proposed RVAE-LAMOL, we conduct our experiments on the three DecaNLP tasks, which is described in Section \ref{section-dataset}, following LAMOL. In addition to multi-task learning, we train on all six permutations of the task order. Then, we obtain the model's performance by calculating the average score on these three tasks. The sampling ratio $\gamma \in [0.01,0.03,0.05,0.2]$, we choose 4 different values to experiment model's performance when pseudo data is lacking and sufficient. As shown in Table \ref{table-deca3}, we can find that RVAE-LAMOL has better performance than LAMOL among all values of $\gamma$. When $\gamma=0.2$, which is the best setting and performance of LAMOL, LAMOL gets 79.5 while RVAE-LAMOL increases by 2.1 percentage point up to 81.6. The performance of RVAE-LAMOL can beat multi-task learning. Compared with LAMOL, the standard deviation of RVAE-LAMOL drops from 0.5 to 0.4, which means RVAE-LAMOL is more robust to the task learning order. \cite{sun2019lamol} has experimented the influence of $\gamma$. The experimental results demonstrate that the performance of LAMOL and the value of $\gamma$ are positively correlated. The larger $\gamma$ leads to better performance while the performance gain disappears when the sampling ratio $\gamma$ is larger than 0.1. Besides, the smaller $\gamma$ makes the model forget how to generate pseudo data of previous tasks resulting in a significant performance decrease.

However, our proposed methods can generate pseudo samples that are more reasonable corresponding to previous tasks when $\gamma$ decreases. When $\gamma=0.05$, RVAE-LAMOL and RCVAE-LAMOL are 2.2 and 3 percentage points higher than LAMOL, respectively. Moreover, the standard deviation of RCVAE-LAMOL and RVAE-LAMOL drops from 1.6 to 1.1. As a decrease of $\gamma$, the advantages of RVAE-LAMOL and RCVAE-LAMOL will be more significant especially RCVAE-LAMOL. When $\gamma=0.03$, both RCVAE-LAMOL and RVAE-LAMOL are 5.8 percentage points higher than LAMOL, respectively. 
\begin{table}[ht]
  \centering
  \caption{Effectiveness of Alternate Lag Training (ALT).}
    \begin{tabular}{cc|cc}
    \toprule
    Modes & Scores & Modes & Scores \\
    \midrule
    $\mathcal{M}_{naive}$ & 75.3  & $\mathcal{M}_{ALT_{M=2}}$ & 79.3 \\
    $\mathcal{M}_{ALT_{M=1}}$ & 74.7  & $\mathcal{M}_{ALT_{M=3}}$ & \textbf{81.3} \\
    $\mathcal{M}_{ALT_{M=1}^{r}}$ & 76.6  & $\mathcal{M}_{ALT_{M=4}}$ & 81.1 \\
    $\mathcal{M}_{ALT_{M=1}^{*}}$ & 80.0    & $\mathcal{M}_{ALT_{M=6}}$ & 80.6 \\
    \bottomrule
    \end{tabular}%
  \label{tab-alt}%
\end{table}The standard deviation drops from 6.2 of LAMOL to 1.6 of RCVAE-LAMOL. When $\gamma=0.01$, RVAE-LAMOL without ID task and RCVAE-LAMOL improves LAMOL by 4.4 and 4.2 percentage points, respectively. In RVAE-LAMOL, the effectiveness of the ID task vanishes when $\gamma=0.01$ because the information of task-specific token [TASK] is not sufficient while RCVAE-LAMOL can supply more task identity information by observing task-specific token [TASK] during LM task and pseudo-data generation. The effectiveness of the ID task and RVAE will be discussed in Section \ref{section effectiveness of id task} and Section \ref{section effectiveness of rvae}, respectively.

\subsection{Effectiveness of Alternate Lag Training (ALT)}\label{section-effect of alt}
In order to validate if ALT is truly helping to train our proposed RVAE-LAMOL and then alleviate catastrophic forgetting, we perform various kinds of training modes. For each mode described below, we only introduce learning one task as an example:
\begin{itemize}
\item $\mathcal{M}_{naive}$: $M=1$. GPT-2 and RVAE are joint trained from start to finish.
\item $\mathcal{M}_{ALT_{M=1}}$: $M=1$. The first $\frac{N_{ep}}{2}$ epochs only train GPT-2, and the remaining epochs joint train GPT-2 and RVAE.
\item $\mathcal{M}_{ALT_{M=1}^{r}}$: $M=1$. The first $\frac{N_{ep}}{2}$ epochs joint train GPT-2 and RVAE, and the remaining epochs only train GPT-2. This mode is a reversed version of $\mathcal{M}_{ALT_{M=1}}$.
\item $\mathcal{M}_{ALT_{M=1}^{*}}$: $M=1$. The first $\frac{N_{ep}}{2}$ epochs only train GPT-2, and the remaining epochs only train RVAE.
\item $\mathcal{M}_{ALT_{M=\eta}}$: $M=\eta, \eta \in [2,3,4,6]$. We split $N_{ep}$ epochs into $\eta$ turns. In each turn, the first $\frac{N_{ep}}{2\eta}$ epochs only train GPT-2, and the remaining epochs only train RVAE.
\end{itemize}
We experiment above training modes on the order SST$\rightarrow$SRL$\rightarrow$WOZ and choose the sampling ratio $\gamma=0.2$ because of the best performance in Section \ref{section deca3 result}. As shown in Table \ref{tab-alt}, we can find that ALT is a better method than naive method $\mathcal{M}_{naive}$ to train RVAE-LAMOL except $\mathcal{M}_{ALT_{M=1}}$. The reason why $\mathcal{M}_{ALT_{M=1}}$ is close to $\mathcal{M}_{naive}$ is that both the encoder and decoder of RVAE are a simple Linear layer, which is easier to be trained than Transformers. Therefore, inconsistent convergence speed (ICS) leads to the unaligned performance of Transformers and RVAE. The similar situation is also occurred in $\mathcal{M}_{ALT_{M=1}^{r}}$. However, $\mathcal{M}_{ALT_{M=1}^{r}}$ is better than $\mathcal{M}_{ALT_{M=1}}$ because the second half of the epochs only train GPT-2, which is much larger than RVAE in terms of parameters, so it can reduce the influence of RVAE.  Since RVAE is trained on the basis of GPT-2 that has been fine-tuned, and there is no problem of ICS, $\mathcal{M}_{ALT_{M=1}^{*}}$ has a significant improvement compared to $\mathcal{M}_{naive}$, $\mathcal{M}_{ALT_{M=1}}$ and $\mathcal{M}_{ALT_{M=1}^{r}}$. However, $\mathcal{M}_{ALT_{M=1}^{*}}$ is a weak improvement than naive LAMOL with the constraint of the simple Linear layer of RVAE. Therefore, we need to consider the problem of ICS and utilize the strong baseline of GPT-2 to improve the training process of RVAE-LAMOL. This is the ALT, $\mathcal{M}_{ALT_{M>2}}$, proposed in this paper. 

As shown in Table \ref{tab-alt}, the advantage of ALT is significant when $M>1$. We split all epochs into $M$ turns. In each turn, we only train GPT-2 in the first half of the turn to obtain a nice baseline. Then, RVAE is jointly trained with the nice GPT-2 in the second half of turn resulting in a certain increase than naive GPT-2. Each turn of increase is gradually accumulated. After ending the M turns training, RVAE-LAMOL is significantly better than other training methods. $M=3$ is a more suitable value to conduct our ALT. $\mathcal{M}_{ALT_{M=3}}$ is 6 percentage points higher than $\mathcal{M}_{naive}$.

\subsection{Effectiveness of the ID task}\label{section effectiveness of id task}
In this section, to validate the effectiveness of the ID task and the influence of the weight of the ID task, we experiment that utilizing only the ID task to extend LAMOL. In order to eliminate the influence of ALT, we used the same naive training $\mathcal{M}_{naive}$ as LAMOL in the experiment and named it as "LAMOL+ID task". As shown in Table \ref{table-deca3}, we can find that the ID task can improve LAMOL in all values of sampling ratio $\gamma$ no matter the order of tasks. When $\gamma=0.2$, the ID task improves LAMOL by 1.1 percentage points with only standard deviation increased by 0.1. When $\gamma=0.05$, the ID task improves LAMOL by 1.2 percentage points meanwhile standard deviation decrease from 1.6 to 0.9. When $\gamma=0.03$, the ID task improves LAMOL by 3.5 percentage points meanwhile standard deviation decrease from 6.2 to 3.5. When $\gamma=0.01$, the ID task improves LAMOL by 1.9 percentage points with only standard deviation increased by 0.4. Surprisingly, the LAMOL+ID task can beat RVAE-LAMOL. We analyze that it is because when the $gamma$ is too small, the wrong task identity information supplied by the ID task results in the training of RVAE biased. However, if the correct task identity information is given to RVAE, as RCVAE, the RVAE can be corrected. The experimental results of RCVAE-LAMOL corroborate our analysis.

\subsection{Effectiveness of Residual Variational Autoencoder (RVAE)}\label{section effectiveness of rvae}
In this section, to validate the effectiveness of RVAE, we experiment that utilizing only the RVAE to improve LAMOL. Firstly, we train RVAE-LAMOL and RCVAE-LAMOL without the ID task same as the experiments in Section \ref{section deca3 result}. Secondly, we explore the influence of the position of RVAE in GPT-2. Finally, we experiment with various dimensions of the latent feature in RVAE to explore the influence of the number of RVAE's parameters.

\begin{figure}[htp]
\centering 
\includegraphics[scale=0.58]{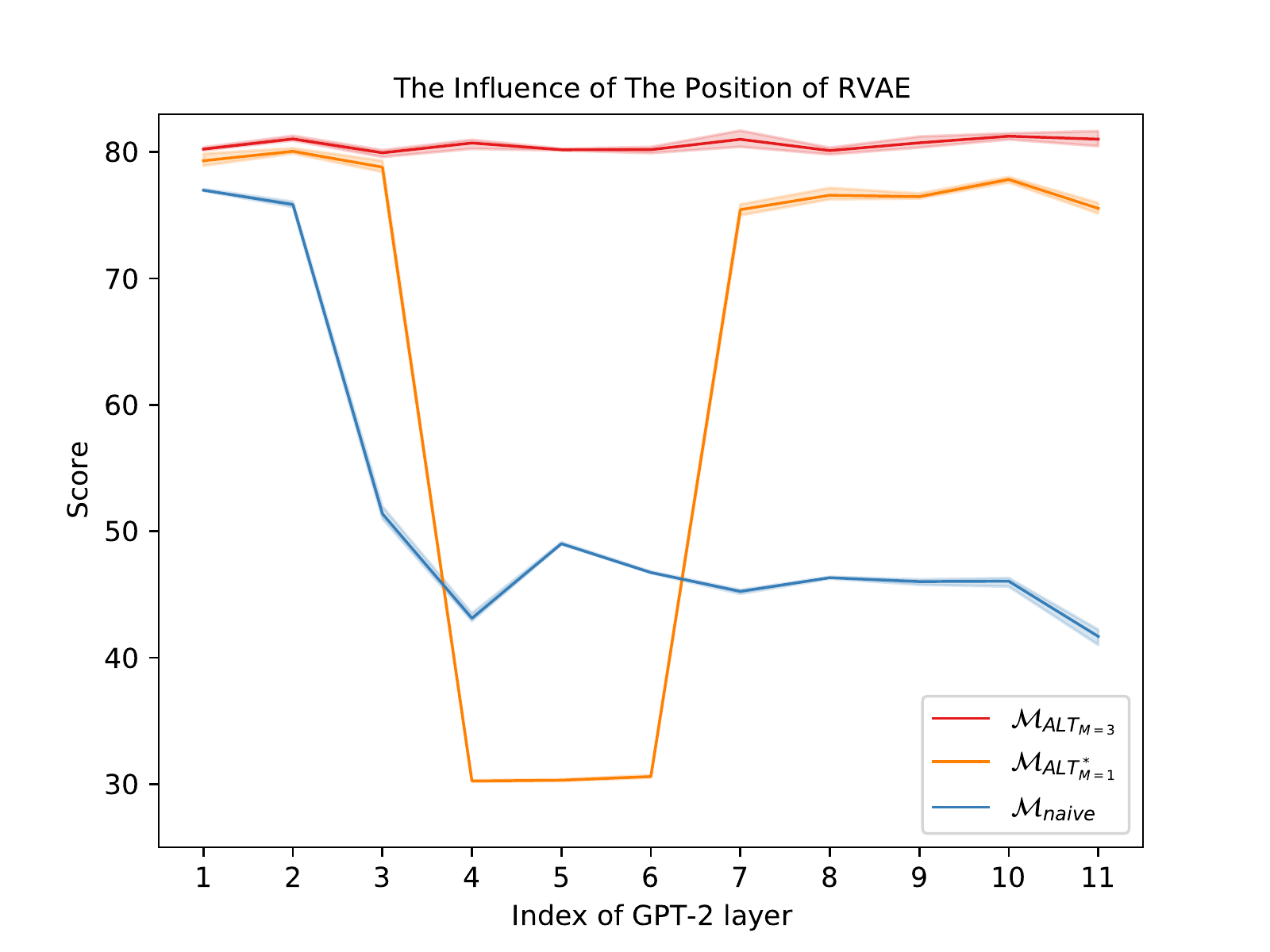} 
\caption{The results of different positions of RVAE in GPT-2.}\vspace{-0.3cm}
\label{fig-vae_idx}
\end{figure}

\subsubsection{Results of RVAE/RCVAE without the ID task}
Same as the experiments of Section \ref{section deca3 result}, we conduct our experiments on four different sampling ratios $\gamma$. As shown in Table \ref{table-deca3}, both RVAE-LAMOL and RCVAE-LAMOL without the ID task perform better than the naive LAMOL and LAMOL with the ID task. When $\gamma=0.2$, RVAE-LAMOL and RCVAE-LAMOL without the ID task improve LAMOL by 1.6 and 1.4 percentage points, respectively. RVAE-LAMOL without the ID task has better performance than RCVAE-LAMOL without the ID task and is more robust to the order of tasks because of lower standard deviation. When the amount of pseudo sample is sufficient, the task identity information of RCVAE has a negative influence. Our analysis is as follows: because the purpose of task identity information is to improve the discrimination between different tasks, if there is too much task identity information, the positive migration between tasks is restrained to some extent because of over discrimination. When $\gamma=0.05$, RVAE-LAMOL and RCVAE-LAMOL without the ID task improve LAMOL by 2.5 and 2.9 percentage points, respectively. 
The standard deviation of LAMOL is decreased from 1.6 to 0.7 and 0.8 with the help of RVAE-LAMOL and RCVAE-LAMOL without the ID task, respectively. Additionally, we can find that RVAE-LAMOL and RCVAE-LAMOL without the ID task become more robust than that with the ID task. When $\gamma=0.03$, RVAE-LAMOL and RCVAE-LAMOL without the ID task improve LAMOL by 5.4 and 5.2 percentage points, respectively. Without the ID task, RVAE-LAMOL decreases a little but is less robust. Without the ID task, RVAE-LAMOL decreases a little but is more robust. When $\gamma=0.01$, RVAE-LAMOL and RCVAE-LAMOL without the ID task improve LAMOL by 4.4 and 5.1 percentage points, respectively. Moreover, RVAE-LAMOL and RCVAE-LAMOL without the ID task are better and more robust than those with ID tasks. We analyze that it is because when the $\gamma$ is too small, the wrong task identity information supplied by the ID task results in the training of RVAE biased.

It can be observed from the above experimental results that RVAE and RCVAE can effectively improve LAMOL. Additionally, the task identity information does not always have a positive effect. The task identity information needs to keep a reasonable proportion in the whole dataset. We will explore how to find a balance between the task identity information and dataset in future work.
\begin{figure}[htp]
\centering 
\includegraphics[scale=0.58]{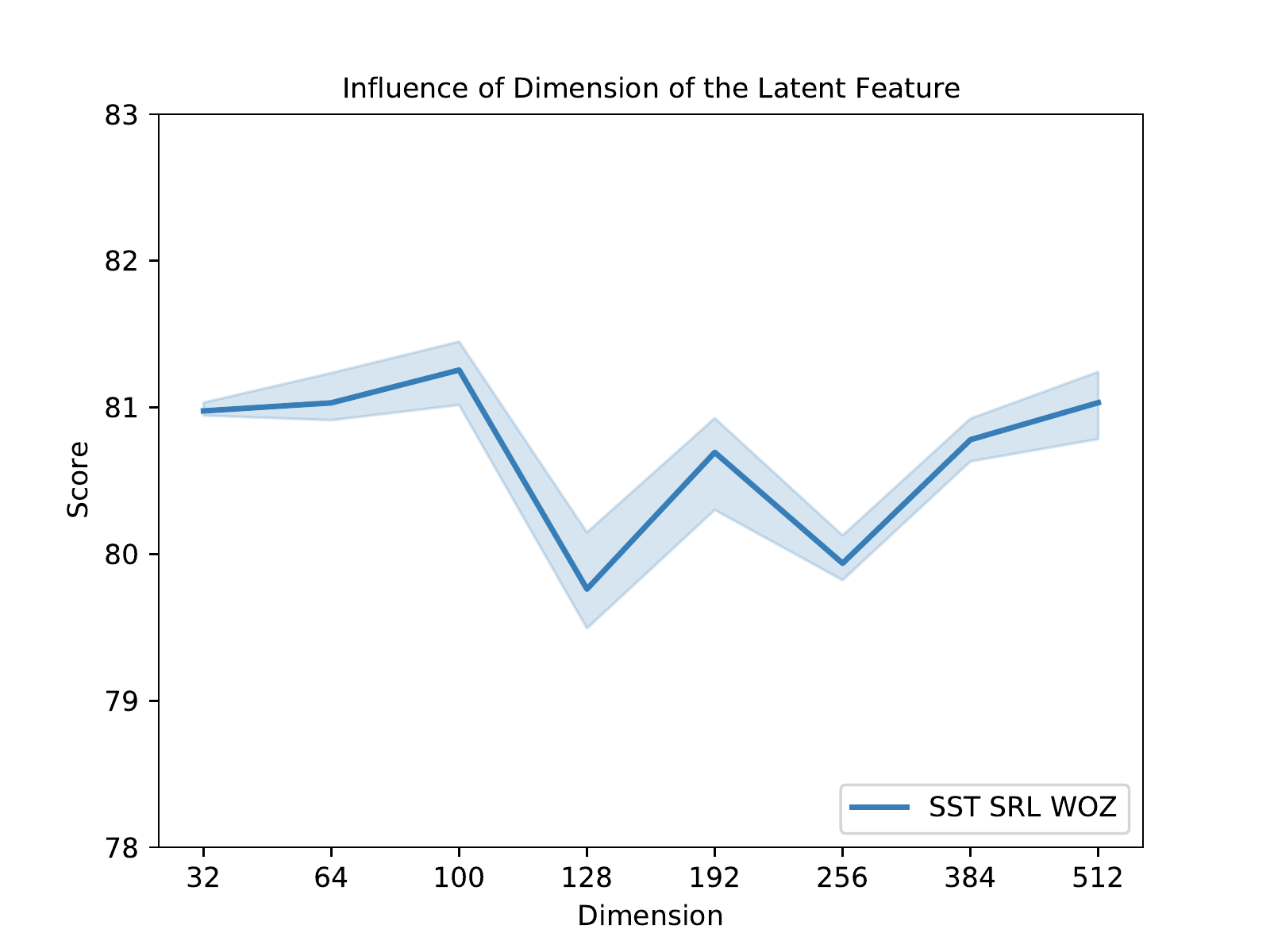} 
\caption{The results of different dimensions of latent feature in RVAE.}\vspace{-0.3cm}
\label{fig-vae_size}
\end{figure}
\subsubsection{The Influence of the Position of RVAE}
According to the results in Table \ref{table-deca3}, we choose RVAE-LAMOL and $\gamma=0.2$ because the best performance. On the task order SST $\rightarrow$ SRL $\rightarrow$ WOZ, we conduct experiments by changing the location of RVAE in GPT-2. Therefore, there are 11 positions for experiments. For each position, we run 3 times and obtain its average and standard deviation. To validate the effectiveness of ALT, we choose the two best training methods, $\mathcal{M}_{ALT_{M=3}}$ and $\mathcal{M}_{ALT_{M=1}^*}$, based on Section \ref{section-effect of alt}. As shown in Fig. \ref{fig-vae_idx}, it demonstrates that our proposed RVAE-LAMOL is robust to the positions in GPT-2. Moreover, $\mathcal{M}_{ALT_{M=3}}$ is essential for training RVAE-LAMOL.

\subsubsection{The Influence of the Dimension of Latent Feature in RVAE}
Since the latent feature is a great component in RVAE, we explore the influence of the dimension of the latent feature in RVAE. On the task order SST $\rightarrow$ SRL $\rightarrow$ WOZ, we change the dimension of the latent feature from small to large and set sampling ratio $\gamma =0.2$ and train with $\mathcal{M}_{ALT_{M=3}}$. For each dimension, we run 3 times and obtain its average and standard deviation. As shown in Fig. \ref{fig-vae_size}, it demonstrates that the performance of RVAE is not proportional to the size of the latent feature. Although a larger latent feature leads to more parameters of RVAE, we can obtain the same or better performance with the small latent feature. To balance computing resources and performance, we set 100 for the dimension of the latent feature in this paper.
\begin{table*}[htbp]
  \centering
  \caption{Examples of pseudo samples from three tasks.}
    \begin{tabular}{m{1cm}<{\centering}|m{3cm}<{\centering}|m{12cm}}
    \toprule
    task  & type  & \multicolumn{1}{c}{content} \\
    \midrule
    \multirow{2}[4]{*}{SST} & corresponding  & \_\_sst\_\_ i liked this movie a lot...is this review negative or positive? \_\_ans\_\_ positive \\
\cmidrule{2-3}          & not corresponding & (similar to QA-SRL) \_\_sst\_\_ after another 2 or 3 days in the fridge the marinated fish is ready to eat.what will be eaten? \_\_ans\_\_ marinated fish\newline{}(similar to WOZ) \_\_sst\_\_ food - - i would like christmas food.what is the change in state? \_\_ans\_\_ food : christmas ; \\
    \midrule
    \multirow{2}[4]{*}{QA-SRL} & corresponding  & \_\_srl\_\_ the first was the open house, which was open to the public on Tuesdays only.\newline{}when was something open to someone? \_\_ans\_\_ on Tuesdays only \\
\cmidrule{2-3}          & not corresponding & (similar to SST) \_\_srl\_\_... too clever by about five tenths.is this review negative or positive? \_\_ans\_\_ negative\newline{}(similar to WOZ) \_\_srl\_\_ i'm looking for a restaurant that serves italian food.what is the change in state? \_\_ans\_\_ food : italian ; \\
    \midrule
    \multirow{2}[4]{*}{WOZ} & corresponding  & \_\_woz.en\_\_ hello, i'm looking for a cheap restaurantwhat is the change in state? \_\_ans\_\_ price range : cheap ; \\
\cmidrule{2-3}          & not corresponding & (similar to SST) \_\_woz.en\_\_ this movie is a total waste of time.is this review negative or positive? \_\_ans\_\_ negative\newline{}(similar to QA-SRL) \_\_woz.en\_\_ there can be no hits on johnson's farm in tg.what can be prevented? \_\_ans\_\_ hit on johnson's farm \\
    \bottomrule
    \end{tabular}%
  \label{tab:pseudo samples}%
\end{table*}%
\section{Conclusion}
To effectively alleviate catastrophic forgetting in LLL for NLP tasks, we propose RVAE-LAMOL, a learning framework that improves the correlation between the sample's content and its own task-specific token by the ID task and RVAE. The ID task tight the correlation by making the model identify which task the content of the sample belongs to. RVAE based on only simple Linear layers can map different tasks into a limited unified semantic space where biased tasks can be corrected by a few pseudo-data. A simple and effective training scheme ALT is proposed to train our proposed RVAE-LAMOL for better performance and more robust to the position of RVAE in GPT-2. Overall, RVAE-LAMOL bridge the gap between LLL and multi-task learning, exhibiting the potential for applying lifelong language learning to true life.

\appendix[Pseudo Samples]
In Table \ref{tab:pseudo samples}, we exhibited some examples of pseudo samples from three tasks in our experiments. The type "\textit{corresponding}" and "\textit{not corresponding}" means that the task-specific tokens, such as "\textit{\_\_sst\_\_}" is the task-specific token of SST, match their corresponding tasks' content or not, respectively.

\section*{Acknowledgements}
This work has been supported by The Youth Innovation Promotion Association of the Chinese Academy of Sciences (E1291902), Jun Zhou (2021025).

\bibliographystyle{IEEEtran}
\bibliography{custom}

\end{document}